\documentclass[sigconf]{acmart}

\usepackage{booktabs} 
\usepackage{bm}
\usepackage{bbding}
\usepackage{amsmath,graphicx}
\usepackage{array}
\usepackage{caption}

\copyrightyear{2017} 
\acmYear{2017} 
\setcopyright{acmcopyright}
\acmConference{MM '17}{October 23--27, 2017}{Mountain View, CA, USA}\acmPrice{15.00}\acmDOI{10.1145/3123266.3123343}
\acmISBN{978-1-4503-4906-2/17/10}

\begin{document}
\title{Single Shot Temporal Action Detection}

\thanks{This research has been supported by the funding from NSFC (61673269, 61273285)  and the Cooperative Medianet Innovation Center (CMIC). * Corresponding author.}

\author{Tianwei Lin$^1$, Xu Zhao$^{1, 3,} $*, Zheng Shou$^2$}
\affiliation{%
  \institution{$^1$Department of Automation, Shanghai Jiao Tong University, China. $^2$Columbia University, USA\\ $^3$Cooperative Medianet Innovation Center (CMIC), Shanghai Jiao Tong University, China\\
  }
}
\email{{wzmsltw,zhaoxu}@sjtu.edu.cn, zs2262@columbia.edu}


\begin{abstract}
Temporal action detection is a very important yet challenging problem, since videos in real applications are usually long, untrimmed and contain multiple action instances. This problem requires not only recognizing action categories but also detecting start time and end time of each action instance. Many state-of-the-art methods adopt the "detection by classification" framework: first do proposal, and then classify proposals. The main drawback of this framework is that the boundaries of action instance proposals have been fixed during the classification step. To address this issue, we propose a novel Single Shot Action Detector (SSAD) network based on 1D temporal convolutional layers to skip the proposal generation step via directly detecting action instances in untrimmed video. On pursuit of designing a particular SSAD network that can work effectively for temporal action detection, we empirically search for the best network architecture of SSAD due to lacking existing models that can be directly adopted. Moreover, we investigate into input feature types and fusion strategies to further improve detection accuracy. We conduct extensive experiments on two challenging datasets: THUMOS 2014 and MEXaction2. When setting Intersection-over-Union threshold to 0.5 during evaluation, SSAD significantly outperforms other state-of-the-art systems by increasing mAP from $19.0\%$ to $24.6\%$  on THUMOS 2014 and from $7.4\%$ to $11.0\%$ on MEXaction2.

\end{abstract}


%
%

\ccsdesc[500]{Computing methodologies~Activity recognition and understanding}


\keywords{Temporal Action Detection, Untrimmed Video, SSAD network}

\maketitle

\begin{figure}
\centering
\setlength{\abovecaptionskip}{-0.05cm} 
\setlength{\belowcaptionskip}{-0.4cm} 
\begin{minipage}[b]{1.0\linewidth}
  \centering
  \centerline{\includegraphics[width=7.6cm]{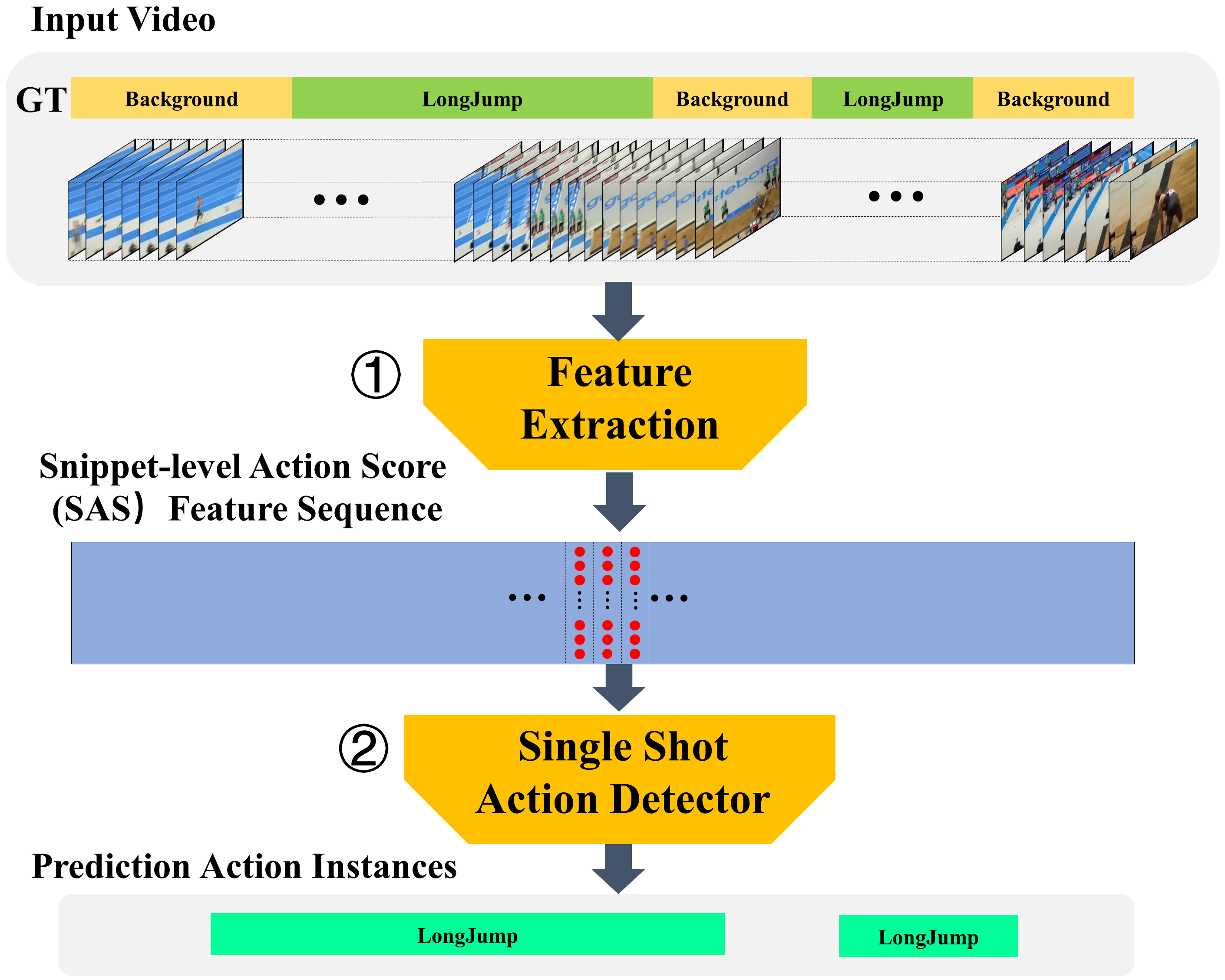}}
  \medskip
\end{minipage}
\caption {Overview of our system. Given an untrimmed long video, (1) we extract Snippet-level Action Score features sequence with multiple action classifiers; (2) SSAD network takes feature sequence as input and directly predicts multiple scales action instances without proposal  generation step.
}
\label{fig_1}
\end{figure}

\section{Introduction}

Due to the continuously booming of videos on the internet, video content analysis has attracted wide attention from both industry and academic field in recently years. An important branch of video content analysis is action recognition, which usually aims at classifying the categories of manually trimmed video clips. Substantial progress has been reported for this task in  \cite{wang2013action,feichtenhofer2016convolutional,wang2015towards,qiu2016deep,tran2015learning}. However, most videos in real world are untrimmed and may contain multiple action instances with irrelevant background scenes or activities. This problem motivates the academic  community to put attention to another challenging task - temporal action detection. This task aims to detect action instances in untrimmed video, including temporal boundaries and categories of instances. Methods proposed for this task can be used in many areas such as surveillance video analysis and intelligent home care.

Temporal action detection can be regarded as a temporal version of object detection in image, since both of the tasks aim to determine the boundaries and categories of multiple instances (actions in time/ objects in space). A popular series of models in object detection are R-CNN and its variants \cite{girshick2015fastt,girshick2014rich,ren2015faster}, which adopt the "detect by classifying region proposals" framework. Inspired by R-CNN, recently many temporal action detection approaches adopt similar framework and classify temporal action instances generated by proposal method \cite{shou2016action,fast_temporal_activity_cvpr16, gangyu,escorcia2016daps} or simple sliding windows method \cite{oneata2014lear,karaman2014fast,wang2014action}. This framework may has some major drawbacks: (1) proposal generation and classification procedures are separate and have to be trained separately, but ideally we want to train them in a joint manner to obtain an optimal model; (2) the proposal generation method or sliding windows method requires additional time consumption; (3) the temporal boundaries of action instances generated by the sliding windows method are usually approximative rather than precise and left to be fixed during classification. Also, since the scales of sliding windows are pre-determined, it is not flexible to predict instances with various scales.

To address these issues, we propose the Single Shot Action Detector (SSAD) network, which is a temporal convolutional network conducted on feature sequence with multiple granularities. Inspired by another set of object detection methods - single shot detection models such as SSD \cite{liu2016ssd} and YOLO \cite{redmon2016you,redmon2016yolo9000}, our SSAD network skips the proposal generation step and directly predicts temporal boundaries and confidence scores for multiple action categories, as shown in Figure \ref{fig_1}. SSAD network contains three sub-modules: (1) base layers read in feature sequence and shorten its temporal length; (2) anchor layers output temporal feature maps, which are associated with anchor action instances; (3) prediction layers generate categories probabilities, location offsets and overlap scores  of these anchor action instances. 


For better encoding of both spatial and temporal information in video, we adopt multiple action recognition models (action classifiers) to extract multiple granularities features. We concatenate the output categories probabilities from all action classifiers in snippet-level and form the Snippet-level Action Score (SAS) feature. The sequences of SAS features  are used as input of SSAD network.

Note that it is non-trivial to adapt the single shot detection model from object detection to temporal action detection. Firstly, unlike VGGNet \cite{Simonyan15}  being used in 2D ConvNet models, there is no existing widely used pre-trained temporal convolutional network.  Thus in this work, we search multiple network architectures to find the best one.
Secondly, we integrate key advantages in  different single shot detection models to make our SSAD network work the best. On one hand, similar to YOLO9000 \cite{redmon2016yolo9000}, we simultaneously predict location offsets, categories probabilities and overlap score of each anchor action instance. On the other hand, like  SSD \cite{liu2016ssd}, we use anchor instances of multiple scale ratios from multiple scales feature maps, which allow network  flexible to handle action instance with various scales. Finally, to further improve performance, we fuse the prediction categories probability with temporal pooled snippet-level action scores during prediction.


The main contributions of our work are summarized as follows:

(1) To the best of our knowledge, our work is the first Single Shot Action Detector (SSAD) for video, which can effectively predict both the boundaries and confidence score of multiple action categories in untrimmed video without the proposal generation step.

(2) In this work, we explore many configurations of SSAD network such as input features type, network architectures and post-processing strategy. Proper configurations  are adopted to achieve better performance for temporal action detection task.

(3) We conduct extensive experiments on two challenging benchmark datasets: THUMOS'14 \cite{jiang2014thumos} and MEXaction2 \cite{mex2}. When setting Intersection-over-Union threshold to 0.5 during evaluation, SSAD significantly outperforms other state-of-the-art systems by increasing mAP from $19.0\%$ to $24.6\%$  on THUMOS'14 and from $7.4\%$ to $11.0\%$ on MEXaction2.

\section{Related Work}
{\bf Action recognition.}
Action recognition is an important research topic for video content analysis. Just as image classification network can be used in image object detection, action recognition models can be used in temporal action detection for feature extraction. We mainly review the following methods which can be used in temporal action detection. Improved Dense Trajectory (iDT) \cite{dtf,wang2013action} feature is consisted of MBH, HOF and HOG features extracted along dense trajectories. iDT method uses SIFT and optical flow to eliminate the influence of camera motion. Two-stream network \cite{feichtenhofer2016convolutional,simonyan2014two,wang2015towards} learns both spatial and temporal features by operating network on single frame and stacked optical flow field respectively using 2D Convolutional Neural Network (CNN) such as GoogleNet \cite{szegedy2015going}, VGGNet \cite{Simonyan15} and ResNet \cite{He_2016_CVPR}. C3D network \cite{tran2015learning} uses 3D convolution to capture both spatial and temporal information directly from raw video frames volume, and is very efficient. Feature encoding methods such as Fisher Vector \cite{wang2013action} and VAE \cite{qiu2016deep} are widely used in action recognition task to improve performance. And there are many widely used action recognition benchmark such as UCF101 \cite{soomro2012ucf101}, HMDB51 \cite{kuehne2013hmdb51} and Sports-1M \cite{sports1m}.

{\bf Temporal action detection.}
This task focuses on learning how to detect action instances in untrimmed videos where the boundaries and categories of action instances have been annotated. Typical datasets such as THUMOS 2014 \cite{jiang2014thumos} and MEXaction2 \cite{mex2} include large amount of untrimmed videos with multiple action categories and complex background information.

Recently, many approaches adopt "detection by classification" framework. For examples, many approaches \cite{oneata2014lear,karaman2014fast,singh2016untrimmed,wang2014action,wang2016uts} use extracted feature such as iDT feature to train SVM classifiers, and then classify the categories of segment proposals or sliding windows using SVM classifiers. And there are some approaches specially proposed for temporal action proposal \cite{fast_temporal_activity_cvpr16, gangyu, escorcia2016daps,apt,spoton_eccv16}. Our SSAD network differs from these methods mainly in containing no proposal generation step.

Recurrent Neural Network (RNN) is widely used in many action detection approaches \cite{yeung2015end,Yuan2016Temporal,ma2016learning,singh2016multi} to encode feature sequence and make per-frame prediction of action categories. However, it is difficult for RNNs to keep a long time period memory in practice \cite{singh2016multi}. An alternative choice is temporal convolution. For example, Lea et al. \cite{lea2016temporal} proposes Temporal Convolutional Networks (TCN) for temporal action segmentation. We also adopt temporal convolutional layers, which makes our SSAD network can handle action instances with a much longer time period.


{\bf Object detection.} 
Deep learning approaches have shown salient performance in object detection. We will review two main set of object detection methods proposed in recent years. The representative methods in first set are R-CNN \cite{girshick2014rich} and its variations \cite{girshick2015fastt,ren2015faster}. R-CNN uses selective search to generate multiple region proposals then apply CNN in these proposals separately to classify their categories; Fast R-CNN \cite{girshick2015fastt} uses a 2D RoI pooling layer which makes feature map  be shared among proposals and reduces the time consumption. Faster RCNN \cite{ren2015faster} adopts a RPN network to generate region proposal instead of selective search. 

Another set of object detection methods are single shot detection methods, which means detecting objects directly without generating proposals. There are two well known models.  YOLO \cite{redmon2016you,redmon2016yolo9000} uses the whole topmost feature map to predict probabilities of multiple categories and corresponding confidence scores and location offsets. SSD \cite{liu2016ssd} makes prediction from multiple feature map with multiple scales default boxes. In our work, we combine the characteristics of these single shot detection methods and embed them into the proposed SSAD network.


\begin{figure*}
\centering
\setlength{\belowcaptionskip}{-0.3cm} 
\begin{minipage}[b]{1.0\linewidth}
  \centering
  \centerline{\includegraphics[width=17cm]{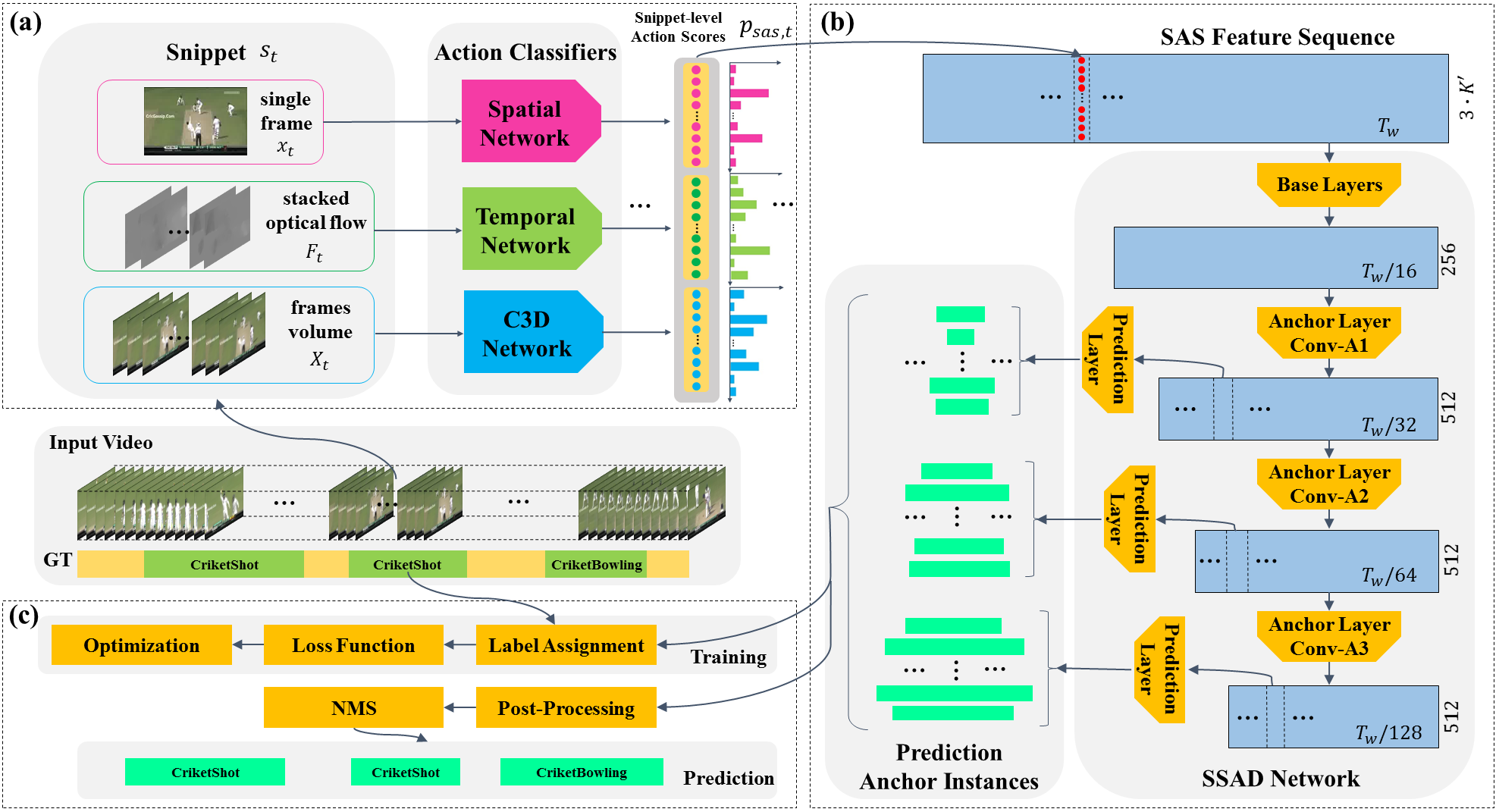}}
  \medskip
\end{minipage}

\caption{The framework of our approach. (a) Multiple action classifiers are used to extract Snippet-level Action Scores (SAS) feature. (b) The architecture of SSAD network: base layers are used to reduce the temporal dimension of input data; anchor layers output multiple scale feature map associated with anchor instances and prediction layers are used for predicting categories, location and confidence of anchor instances. (c) The training and prediction procedures: during training, we match anchor instances with ground truth instances and calculate loss function for optimization. During prediction, post-processing and NMS procedure are conducted on anchor instances to make final prediction.
}
\label{fig_2}

\end{figure*}

\section{Our Approach}


In this section, we will introduce our approach in details. The framework of our approach is shown in Figure \ref{fig_2}. 

\subsection{Problem Definition}

We denote a video as $X_v=\left \{ x_t \right \}_{t=1}^{T_v}$ where $T_v$ is the number of frames in $X_v$ and $x_t$ is the $t$-th frame in $X_v$. Each untrimmed video $X_v$ is annotated with a set of temporal action instances $\Phi _v = \left \{ \phi _n=\left (\varphi _n,\varphi' _n, k_n   \right ) \right \}_{n=1}^{N_v}$, where $N_v$ is the number of temporal action instances in $X_v$, and $\varphi _n,\varphi' _n, k_n $ are starting time, ending time and category of action instance $\phi_n$ respectively. $k_n \in \left \{1,...,K  \right \} $ where $K$ is the number of action categories. $\Phi _v$ is given during training procedure and need to be predicted during prediction procedure.

\subsection{Extracting of Snippet-level Action Scores}
To apply SSAD model, first we need to make snippet-level action classification and get Snippet-level Action Score (SAS) features. Given a video $X_v$, a snippet $s_t=\left ( x_t, F_t, X_t \right )$ is composed by three parts: $x_t$ is the $t$-th frame in $X_v$, $F_t=\left \{f_{t'}  \right \}_{t'=t-4}^{t+5}$ is stacked optical flow field derived around $x_t$ and $X_t=\left \{ x_{t'} \right \}_{t'=t-7}^{t+8}$ is video frames volume. So given a video $X_v$, we can get a sequence of snippets $S_v=\left \{ s_t \right \}_{t=1}^{T_v}$. We pad the video $X_v$ in head and tail with first and last frame separately to make $S_v$ have the same length as $X_v$.

{\bf Action classifier.} To evaluate categories probability of each snippet, we use multiple action classifiers with commendable performance in action recognition task: two-stream network \cite{simonyan2014two} and C3D network \cite{tran2015learning}. Two-stream network includes spatial and temporal networks which operate on single video frame $x_t$ and stacked optical flow field $F_t$ respectively. We use the same two-stream network architecture as described in \cite{wang2015towards}, which adopts VGGNet-16 network architecture. 
C3D network is proposed in \cite{tran2015learning}, including multiple 3D convolution layers and 3D pooling layers. C3D network operates on short video frames volume $X_t$ with length $l$, where $l$ is the length of video clip and is set to 16 in C3D. So there are totally three individual action classifiers, in which spatial network measures the spatial information, temporal network measures temporal consistency and C3D network measures both. In section 4.3, we evaluate the effect of each action classifier and their combinations.

{\bf SAS feature.} As shown in Figure \ref{fig_2}(a), given a snippet $s_t$, each action classifier can generate a score vector $\bm{p_t}$ with length $K'=K+1$, where $K'$ includes $K$ action categories and one background category. Then we concatenate output  scores of each classifiers to form the {\bf Snippet-level Action Score} (SAS) feature $\bm{p_{sas,t}}=\left ( \bm{p_{S,t}},\bm{ p_{T,t}}, \bm{p_{C,t}} \right )$, where $\bm{p_{S,t}}$, $\bm{ p_{T,t}}$, $\bm{p_{C,t}} $ are output score of spatial, temporal and C3D network separately.  So given a snippets sequence $S_v$ with length $T_v$, we can extract a SAS feature sequence $P_v=\left \{ \bm{p_{sas,t}} \right \}_{t=1}^{T_v}$. Since the number of frames in video is uncertain and may be very large, we use a large observation window with length $T_w$ to truncate the feature sequence. We denote a window as $\omega =\left \{ \varphi _{\omega},\varphi '_{\omega} ,P_{\omega},\Phi _{\omega}  \right \} $, where $\varphi _{\omega}$ and $\varphi '_{\omega}$ are starting and ending time of $\omega$, $P_{\omega}$ and $\Phi _{\omega}$ are SAS feature sequence and corresponding ground truth action instances separately. 


\subsection{SSAD Network}
Temporal action detection is quite different from object detection in 2D image. In SSAD we adopt two main characteristics from single shot object detection models such as SSD \cite{liu2016ssd} and YOLO \cite{redmon2016you,redmon2016yolo9000}: 1) unlike "detection by classification" approaches, SSAD directly predicts  categories and location offsets of action instances in untrimmed video using convolutional  prediction layers; 2) SSAD combine temporal feature maps from different convolution layers for prediction, making it possible to handle action instances with various length. We first introduce the network architecture.



{\bf Network architecture.}
The architecture of SSAD network is presented in Figure \ref{fig_2}(b), which mainly contains three sub-modules: base layers, anchor layers and prediction layers. Base layers handle the input SAS feature sequence, and use both convolution and pooling layer to shorten the temporal length of feature map and increase the size of receptive fields. Then anchor layers use temporal convolution to continually shorten the feature map and output anchor feature map for action instances prediction. Each cell of anchor layers is associated with anchor instances of multiple scales. Finally, we use  prediction layers to get classification score, overlap score and location offsets of each anchor instance.

In SSAD network, we adopt 1D temporal convolution and pooling to capture temporal information. We conduct Rectified Linear Units (ReLu) activation function \cite{glorot2011deep} to output temporal feature map except for the convolutional prediction layers. And we adopt temporal max pooling since max pooling can enhance the invariance of small input change.

\begin{figure}
\centering
\begin{minipage}[b]{1.0\linewidth}
  \centering
  \centerline{\includegraphics[width=8.4cm]{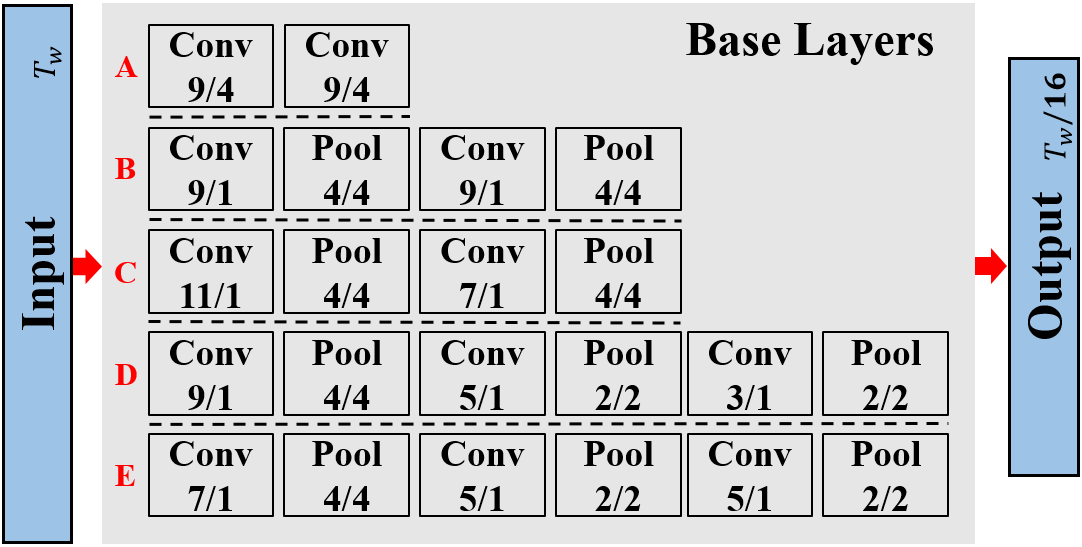}}
  \medskip
\end{minipage}
\caption{Multiple architectures of base layers. Input and output sizes are same for each architecture. Parameter of layer is shown with the format of $kernel/stride$. All convolutional layers have 512 convolutional filters. Evaluation results of these architectures are shown in section 4.3, and we adopt architecture $B$ which achieves the best performance.}
\label{fig_bn}
\end{figure}

{\bf Base layers.}
Since there are no widely used pre-trained 1D ConvNet models such as the VGGNet \cite{Simonyan15} used in 2D ConvNet models, we search many different network architectures for SSAD network. These architectures only differ in base layers while we keep same architecture of anchor layers and prediction layers. As shown in Figure \ref{fig_bn}, we totally design 5 architectures of base layers. In these architectures, we mainly explore three aspects: 1) whether use convolution or pooling layer to shorten the temporal dimension and increase the size of receptive fields; 2) number of layers of network and 3) size of convolution layer's kernel.  Notice that we set the number of convolutional filter in all base layers to 256. Evaluation results of these architectures are shown in section 4.3, and finally we adopt architecture $B$ which achieves the best performance.


{\bf Multi-scale anchor layers.}
After processing SAS feature sequence using base layers, we stack three anchor convolutional layers (Conv-A1, Conv-A2 and Conv-A3) on them. These layers have same configuration: kernel size 3, stride  size 2 and 512 convolutional filters. The output anchor feature maps of anchor layers are $f_{A1}$, $f_{A2}$ and $f_{A3}$ with size $(T_w/32\times 512)$, $(T_w/64\times 512)$ and $(T_w/128\times 512)$ separately. Multiple anchor layers decrease temporal dimension of feature map progressively and allow SSAD get predictions from multiple resolution feature map. 

\begin{figure}
\centering
\begin{minipage}[b]{1.0\linewidth}
  \centering
  \centerline{\includegraphics[width=8.2cm]{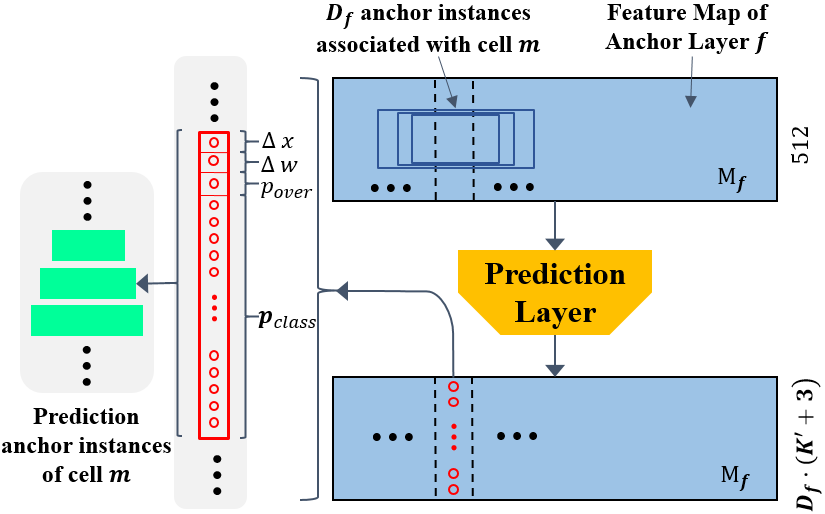}}
  \medskip
\end{minipage}
\caption{Anchor instances and prediction layer in temporal feature map. In feature map of a anchor layer, we associate a set of multiple scale anchor instances with each feature map cell. We use convolutional prediction layer to predict location offset, confidence and classification scores simultaneously for each anchor instance.
}
\label{fig_3}
\end{figure}

For each temporal feature map of anchor layers, we associate a set of multiple scale anchor action instances with each feature map cell as shown in Figure \ref{fig_3}. For each anchor instance, we use convolutional prediction layers to predict  overlap score, classification score and location offsets, which will be introduced later. 

In term of the details of multi-scale anchor instances, the lower anchor feature map has higher resolution and smaller receptive field than the top anchor feature map. So we let the lower anchor layers detect short action instances and the top anchor layers detect long action instances. For a temporal feature map $f$ of anchor layer with length $M$, we define base scale $s_f=\frac{1}{M}$ and a set of scale ratios $R_f= \left \{r_d \right \}_{d=1}^{D_f} $, where $D_f$ is the number of scale ratios. We use $ \{ 1,1.5,2 \}$ for $f_{A1}$ and $ \{0.5,0.75, 1,1.5,2 \}$ for $f_{A2}$ and $f_{A3}$. For each ratio $r_d$, we calculate $\mu _w=s_f\cdot r_d $ as anchor instance's default width. And all anchor instances associated with the $m$-th feature map cell share the same default center location $\mu _c=\frac{m+0.5}{M}$. So for an anchor feature map $f$ with length $M_f$ and $D_f$ scale ratios, the number of associated anchor instances is $M_f\cdot  D_f$.

{\bf Prediction layers.}
We use a set of convolutional filters to predict classification scores, overlap scores and location offsets of anchor instances associated with each feature map cell. As shown in Figure \ref{fig_3}, for an anchor feature map $f$ with length $M_f$ and $D_f$ scale ratios, we use $D_f\cdot (K'+3)$ temporal convolutional filters with kernel size 3, stride size 1 for prediction. The output of prediction layer has size $\left (M_f\times \left ( D_f\cdot (K'+3) \right )   \right ) $  and can be reshaped into $\left (\left (M_f\cdot  D_f \right ) \times (K'+3)  \right)$. Each anchor instance gets a prediction score vector $\bm{p_{pred}}=\left (\bm{p_{class}},p_{over},\Delta c,\Delta w  \right )$ with length $(K'+3)$, where $\bm{p_{class}}$ is classification score vector with length $K'$, $p_{over}$ is overlap score and $\Delta c$, $\Delta w$ are location offsets. Classification score $p_{class}$ is used to predict anchor instance's category. Overlap score $p_{over}$ is used to estimate the overlap between anchor instance and ground truth instances and should have value between $[0,1]$, so it is normalized by using sigmoid function:

\begin{equation}
p'_{over}=sigmoid(p_{over}){.}
\end{equation}

And location offsets $\Delta c$, $\Delta w$ are used for adjusting the default location of anchor instance. The adjusted location is defined as:

\begin{equation}
\begin{split}
& \varphi_c=\mu _c+\alpha _1 \cdot \mu _w \cdot \Delta c \\
& \varphi_w=\mu _w\cdot exp(\alpha_2 \cdot\Delta w){,} \\
\end{split}
\end{equation}
where $\varphi_c$ and $\varphi_w$ are center location and width of anchor instance respectively. $\alpha _1$ and $\alpha _2$ are used for controlling the effect of location offsets to make prediction stable.  We set both $\alpha _1$ and $\alpha _2$ to 0.1. The starting and ending time of action instance are $\varphi=\varphi_c- \frac{1}{2}\cdot \varphi_w$ and $\varphi'=\varphi_c+ \frac{1}{2}\cdot \varphi_w$ respectively. So for a anchor feature map $f$, we can get a anchor instances set $\Phi _f = \left \{ \phi _n=\left (\varphi _c,\varphi _w, p_{class},p'_{over}   \right ) \right \}_{n=1}^{N_f}$, where $N_f=M_f\cdot  D_f$ is the number of anchor instances. And the total prediction instances set is $\Phi _p=\left \{\Phi _{f_{A1}},\Phi _{f_{A2}} ,\Phi _{f_{A3}}   \right \}$.

\subsection{Training of SSAD network}

{\bf Training data construction.}
As described in Section 3.2, for an untrimmed video $X_v$ with length $T_v$, we get SAS features sequence $P_v$ with same length. Then we slide window of length $T_w$ in feature sequence with $75\%$ overlap. The overlap of sliding window is aim to handle the situation where action instances locate in boundary of window and also used to increase the amount of training data. During training, we only keep windows containing at least one ground-truth instance. So given a set of untrimmed training videos, we get a training set  $\Omega= \left \{ \omega_n \right \}_{n=1}^{N_{\omega }}$, where $N_{\omega}$ is the number of windows. We randomly shuffle the data order in training set to make the network converge faster, where same random seed is used during evaluation.

{\bf Label assignment.}
During training, given a window $\omega$, we can get  prediction instances set $\Phi _p$ via SSAD network. We need to match them with ground truth set $\Phi _{\omega} $ for label assignment. For an anchor instance $\phi_n$ in $\Phi _p $, we calculate it's IoU overlap with all ground truth instances in $\Phi _{\omega} $. If the highest IoU overlap is higher than 0.5, we match $\phi_n$ with corresponding ground truth instance $\phi_g$ and regard it as positive, otherwise negative. We expand $\phi_n$ with matching information as $\phi' _n=\left (\varphi _c,\varphi _w, \bm{p_{class}}, p'_{over},k_g,g_{iou},g_c, g_w   \right )$, where $k_g$ is the category of $\phi_g$ and is set to 0 for negative instance, $g_{iou}$ is the IoU overlap between $\phi_n$ and $\phi_g$, $g_c$ and $g_w $ are center location and width of $\phi_g$ respectively. So a ground truth instance can match multiple anchor instances while a anchor instance can only match one ground truth instance at most.

{\bf Hard negative mining.} 
During label assignment, only a small part of anchor instances match the ground truth instances, causing an imbalanced data ratio between the positive and negative instances. Thus we adopt the hard negative mining strategy to reduce the number of negative instances. Here, the hard negative instances are defined as negative instances with larger overlap score than 0.5. We take all hard negative instances and randomly sampled negative instances in remaining part to make the ratio between positive and negative instances be nearly 1:1. This ratio is chosen by empirical validation. So after label assignment and hard negative mining, we get $\Phi'_p=\left \{\phi'_n  \right \}_{n=1}^{N_{train}}$ as the input set during training, where $N_{train}$ is the number of total training instances and is the sum of the number of positives $N_{pos}$ and negatives $N_{neg}$.



{\bf Objective for training.}
The training objective of the SSAD network is to solve a multi-task optimization problem. The overall loss function is a weighted sum of the classification loss (class), the overlap loss (conf), the detection loss (loc) and L2 loss for regularization:

\begin{equation}
L = L_{class}+\alpha \cdot L_{over} +\beta \cdot  L_{loc}+ \lambda \cdot L_2(\Theta),
\end{equation}
where $\alpha $, $\beta $ and $\lambda $ are the weight terms used for balancing each part of loss function. Both $\alpha $ and $\beta $ are set to 10 and $\lambda $ is set to 0.0001 by empirical validation. For the classification loss, we use conventional softmax loss over multiple categories, which is effective for training classification model and can be defined as:

\begin{equation}
L_{class} = L_{softmax}=\frac{1}{N_{train}}\sum_{i=1}^{N_{train}}(-log(P_{i}^{(k_g)})),
\end{equation}
where $P_{i}^{(k_g)}=\frac{exp(p_{class,i}^{(k_g)})}{\sum_j exp(p_{class,i}^{(k_j)})}$ and $k_g$ is the label of this instance.

$L_{over}$ is used to make a precise prediction of anchor instances' overlap IoU score, which helps the procedure of NMS. The overlap loss adopts the mean square error (MSE) loss and be defined as:

\begin{equation}
L_{over} =\frac{1}{N_{train}}\sum_{i=1}^{N_{train}}(p'_{over,i}-g_{iou,i}).
\end{equation}

$L_{loc}$ is the Smooth L1 loss \cite{girshick2015fastt} for location offsets. We regress the center ($\phi_c$) and width ($\phi_w$) of predicted instance:

\begin{equation}
L_{loc} =\frac{1}{N_{pos}}\sum_{i=1}^{N_{pos}}(SL_1(\phi_{c,i}-g_{c,i})+SL_1(\phi_{w,i}-g_{w,i})),
\end{equation}
where $g_{c,i}$ and $g_{w,i}$ is the center location and width of ground truth instance. $L_2(\Theta)$ is the L2 regularization loss where $\Theta$ stands for the parameter of the whole SSAD network.

\subsection{Prediction and post-processing}
During prediction, we follow the aforementioned data preparation method during the training procedure to prepare test data, with the following two changes: (1) the overlap ratio of window is reduced to $25\%$ to increase the prediction speed and reduce the redundant predictions; (2) instead of removing windows without annotation, we keep all windows during prediction because the removing operation is actually a leak of annotation information. If the length of input video is shorter than $T_w$, we will pad SAS feature sequence to $T_w$ so that there is at least one window for prediction. Given a video $X_v$, we can get a set of $\Omega= \left \{ \omega_n \right \}_{n=1}^{N_{\omega }}$. Then we use SSAD network to get prediction anchors of each window and merge these prediction as $\Phi_p=\left \{\phi_n   \right \}_{n=1}^{N_p}$, where ${N_p}$ is the number of prediction instances. For a prediction anchor instance $\phi_n$ in $\Phi_p$, we calculate the mean Snippet-level Action Score $\bm{\bar{p}}_{sas}$ among the temporal range of instance and multiple action classifiers. 

\begin{equation}
\bm{\bar{p}}_{sas}=\frac{1}{3\cdot (\varphi'-\varphi  )}\sum_{t=\varphi }^{\varphi '}\left ( \bm{p}_{S,t}+ \bm{ p}_{T,t}+ \bm{p}_{C,t} \right ),
\end{equation}
where $\varphi $ and $\varphi'$ are starting and ending time of prediction anchor instance respectively. Then we fuse categories scores $\bm{\bar{p}}_{sas}$ and $\bm{p}_{class}$ with multiplication factor $p_{conf}$ and get the $\bm{p_{final}}$:

\begin{equation}
\bm{p}_{final}=p'_{over}\cdot \left ( \bm{p}_{class} +\bm{\bar{p}}_{sas} \right ).
\end{equation}

We choose the maximum dimension $k_p$ in $\bm{p}_{final}$ as the category of $\phi_n$ and corresponding score $p_{conf}$ as the confidence score. We expand $\phi_n$ as $\phi'_n=\left \{\varphi _c,\varphi _w, p_{conf},k_p  \right \}$ and get prediction set $\Phi' _p=\left \{\phi'_n   \right \}_{n=1}^{N_p}$. Then we conduct non-maximum suppress (NMS) in these prediction results to remove redundant predictions with confidence score $p_{conf}$ and get the final prediction instances set $\Phi'' _p=\left \{\phi'_n   \right \}_{n=1}^{N_{p'}}$, where $N_{p'}$ is the number of the final prediction anchors. Since there are little overlap between action instances of same category in temporal action detection task, we take a strict threshold in NMS, which is set to 0.1 by empirical validation. 


\section{Experiments}
\subsection{Dataset and setup}
{\bf THUMOS 2014 \cite{jiang2014thumos}.} 
The temporal action detection task of THUMOS 2014 dataset is challenging and widely used. The training set is the UCF-101 \cite{soomro2012ucf101} dataset including 13320 trimmed videos of 101 categories. The validation and test set contain 1010 and 1574 untrimmed videos separately. In temporal action detection task, only 20 action categories are involved and annotated temporally. We only use 200 validation set videos (including 3007 action instances) and 213 test set videos (including 3358 action instances) with temporal annotation to train and evaluate SSAD network.

{\bf MEXaction2 \cite{mex2}.} There are two action categories in MEXaction2 dataset: "HorseRiding" and "BullChargeCape". This dataset is consisted of three subsets: YouTube clips, UCF101 Horse Riding clips and INA videos. YouTube and UCF101 Horse Riding clips are trimmed and used for training set, whereas INA videos are untrimmed with approximately 77 hours in total and are divided into training, validation and testing set. Regarding to temporal annotated action instances, there are 1336 instances in training set, 310 instances in validation set and 329 instances in testing set.

{\bf Evaluation metrics.} 
For both datasets, we follow the conventional metrics used in THUMOS'14, which evaluate Average Precision (AP) for each action categories and calculate mean Average Precision (mAP) for evaluation. A prediction instance is correct if it gets same category as ground truth instance and its temporal IoU with this ground truth instance is larger than IoU threshold $\theta$. Various IoU thresholds are used during evaluation. Furthermore, redundant detections for the same ground truth are forbidden.


\subsection{Implementation Details}
{\bf Action classifiers.} 
To extract SAS features, action classifiers should be trained first, including two-stream networks \cite{wang2015towards} and C3D network \cite{tran2015learning}. We implement both networks based on Caffe \cite{jia2014caffe}. For both MEXaction and THUMOS'14 datasets, we use trimmed videos in training set to train action classifier.

For spatial and temporal network, we follow the same training strategy described in \cite{wang2015towards} which uses the VGGNet-16 pre-trained on ImageNet \cite{deng2009imagenet} to intialize the network and fine-tunes it on training set. And we follow \cite{tran2015learning} to train the C3D network, which is pre-trained on Sports-1M \cite{sports1m} and then is fine-turned on training set. 


\begin{figure*}
\setlength{\abovecaptionskip}{0cm}
\begin{minipage}[b]{1.0\linewidth}
  \centering
  \centerline{\includegraphics[width=17.3cm]{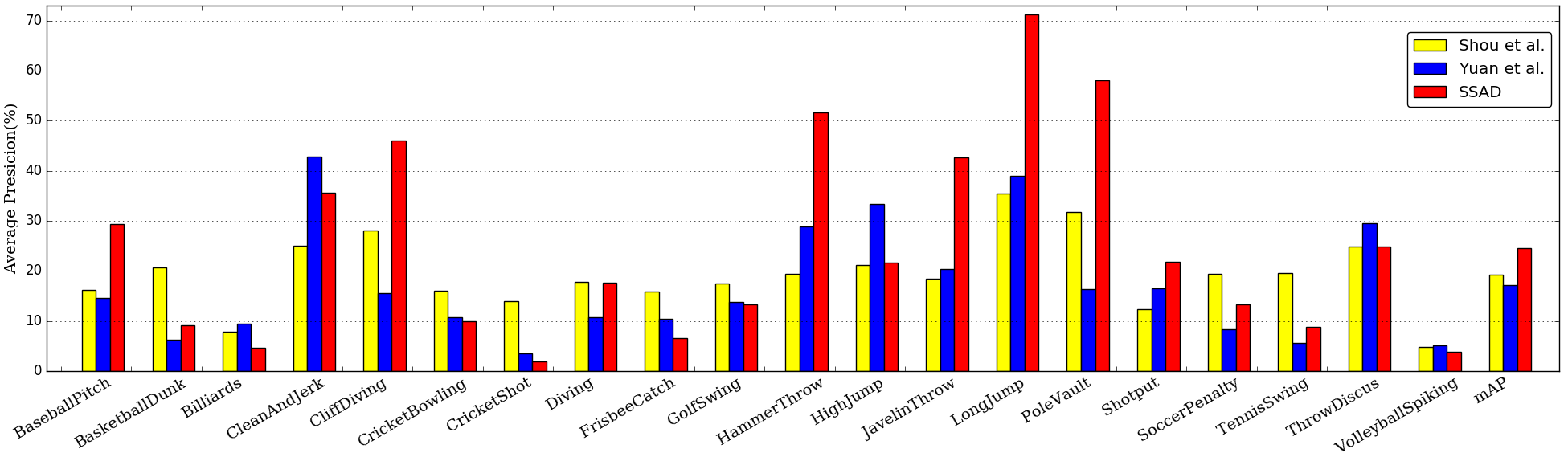}}
  \medskip
\end{minipage}
\caption{Detection AP over different action categories with overlap threshold 0.5 in THUMOS'14.}
\label{fig_result}
\end{figure*}

{\bf SSAD optimization.} 
For training of the SSAD network, we use the adaptive moment estimation (Adam) algorithm \cite{kingma2014adam} with the aforementioned multi-task loss function. Our implementation is based on Tensorflow \cite{abadi2016tensorflow}. We adopt the Xavier method \cite{glorot2010understanding} to  randomly initialize parameters of whole SSAD network because there are no suitable pre-trained temporal convolutional network. Even so, the SSAD network can be easily trained with quick convergence since it has a small amount of parameters (20 MB totally) and the input of SSAD network - SAS features are concise high-level feature. The training procedure takes nearly 1 hour on THUMOS'14 dataset.


\begin{table}[!htbp]
\centering
\caption{mAP results on THUMOS'14 with various IoU threshold $\theta $ used in evaluation.}
\begin{tabular}{p{2.5cm}<{\centering}p{0.75cm}<{\centering}p{0.75cm}<{\centering}p{0.75cm}<{\centering}p{0.75cm}<{\centering}p{0.75cm}<{\centering}}
\toprule
 $\theta $ & 0.5 & 0.4 & 0.3 & 0.2 & 0.1 \\
\midrule  
Karaman et al. \cite{karaman2014fast} & 0.2 & 0.3 & 0.5 & 0.9 & 1.5 \\
Wang et al. \cite{wang2014action} & 8.5 & 12.1 & 14.6 & 17.8 & 19.2\\
Oneata et al. \cite{oneata2014lear} & 15.0 & 21.8 & 28.8 & 36.2 & 39.8\\
Richard et al. \cite{Richard_2016_CVPR}  & 15.2 & 23.2 & 30.0 & 35.7 & 39.7\\
Yeung et al. \cite{yeung2015end} & 17.1 & 26.4 & 36.0 & 44.0 & 48.9\\
Yuan et al. \cite{Yuan2016Temporal} & 18.8 & 26.1 & 33.6 & 42.6 &  {\bf 51.4}\\
Shou et al. \cite{shou2016action} &  19.0 & 28.7 & 36.3 &  43.5 & 47.7\\
Zhu et al. \cite{zhu2016efficient} & 19.0 & 28.9 & 36.2 &  43.6 & 47.7\\
\midrule SSAD & {\bf 24.6} & {\bf 35.0} & {\bf 43.0} & {\bf 47.8} &  50.1\\
\bottomrule
\end{tabular}

\label{table_thumos}

\end{table}

\subsection{Comparison with state-of-the-art systems}

{\bf Results on THUMOS 2014.}
To train action classifiers, we use full UCF-101 dataset. Instead of using one background category, here we form background categories using 81 action categories which are un-annotated in detection task. Using two-stream and C3D networks as action classifiers, the dimension of SAS features is 303.

For training of SSAD model, we use 200 annotated untrimmed video in THUMOS'14 validation set as training set. The window length $L_w$ is set to 512, which means approximately 20 seconds of video with 25 fps. This choice is based on the fact that $99.3\%$ action instances in the training set have smaller length than 20 seconds. We train SSAD network for 30 epochs with learning rate of 0.0001.

\begin{figure*}
\centering

\begin{minipage}[b]{1.0\linewidth}
  \centering
  \centerline{\includegraphics[width=17.7cm]{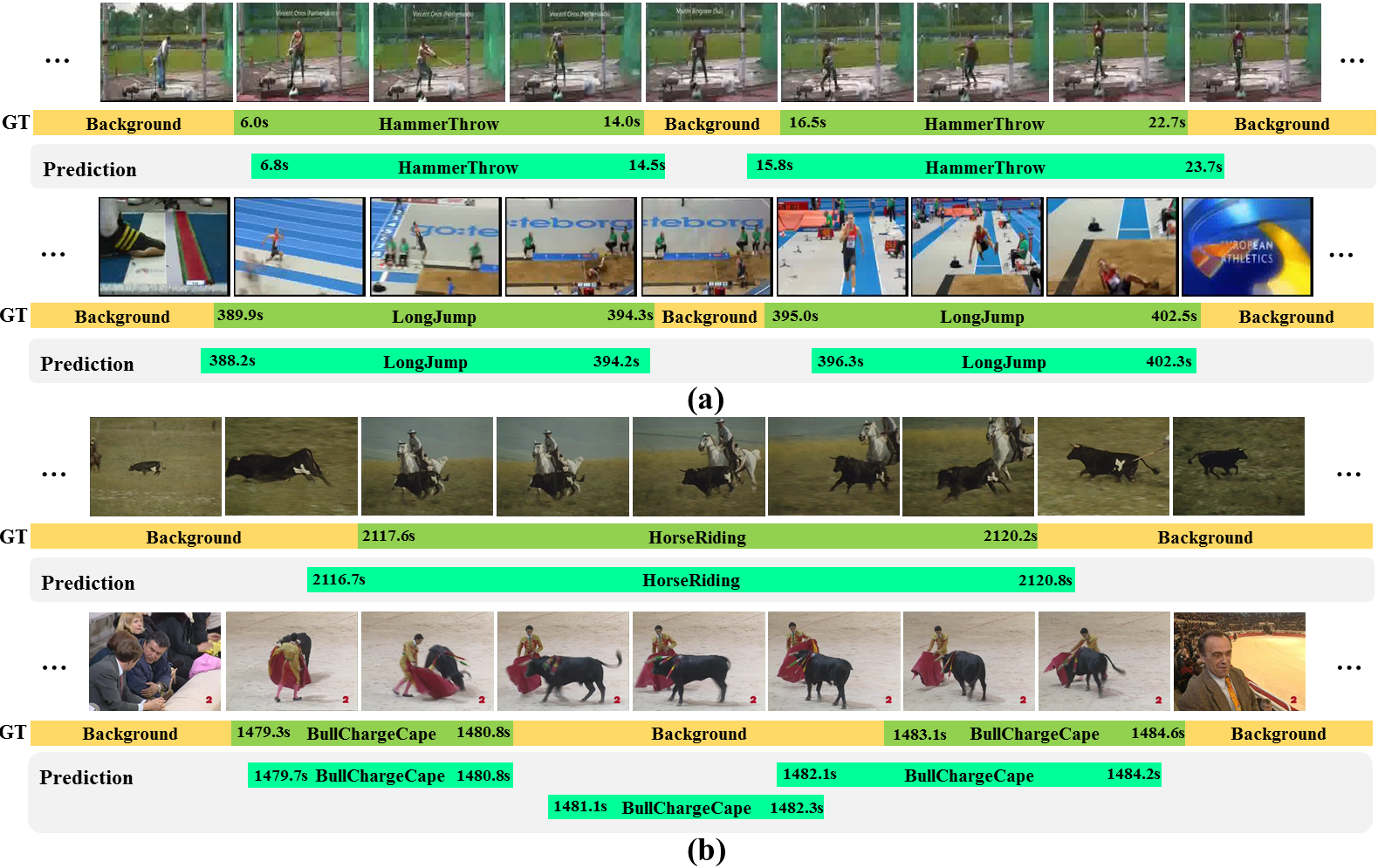}}
  \medskip
\end{minipage}

\caption{Visualization of prediction action instances by SSAD network. Figure (a) shows prediction results for two action categories in THUMOS'14 dataset. Figure (b) shows prediction results for two action categories in MEXaction2 dataset.
}
\label{fig_vs}
\end{figure*}

The comparison results between our SSAD and other state-of-the-art systems are shown in Table \ref{table_thumos} with multiple overlap IoU thresholds varied from 0.1 to 0.5. These results show that SSAD significantly outperforms the compared state-of-the-art methods. While the IoU threshold used in evaluation is set to 0.5, our SSAD network improves the state-of-the-art mAP result from $19.0\%$ to $24.6\%$. The Average Precision (AP) results of all categories with overlap threshold 0.5 are shown in Figure \ref{fig_result}, the SSAD network outperforms other state-of-the-art methods for 7 out of 20 action categories. Qualitative results are shown in Figure \ref{fig_vs}.

\begin{table}[!tbp]
\centering
\caption{Results on MEXaction2 dataset with overlap threshold 0.5. Results for \cite{mex2} are taken from \cite{shou2016action}.}
\begin{tabular}{p{1.8cm}<{\centering}ccp{1.5cm}<{\centering}}
\toprule
AP$(\%)$ & BullCHargeCape & HorseRiding & mAP$(\%)$  \\
\midrule DTF \cite{mex2}  & 0.3 & 3.1 & 1.7 \\
SCNN \cite{shou2016action}  & 11.6 & 3.1 & 7.4 \\
\midrule SSAD & {\bf 16.5} & {\bf 5.5} & {\bf 11.0} \\
\bottomrule
\end{tabular}
\label{result_mex}
\end{table}

{\bf Results on MEXaction2.}
For training of action classifiers, we use all 1336 trimmed video clips in training set. And we randomly sample 1300 background video clips in untrimmed training videos. The prediction categories of action classifiers are "HorseRiding", "BullChargeCape" and "Background". So the dimension of SAS features equals to 9 in MEXaction2.

For SSAD model, we use all 38 untrimmed video in MEXaction2 training set training set. Since the distribution of action instances' length in MEXaction2 is similar with THUMOS'14, we also set the interval of snippets to zero and the window length $T_w$ to 512. We train all layers of SSAD for 10 epochs with learning rate of 0.0001. 

We compare SSAD with SCNN \cite{shou2016action} and typical dense trajectory features (DTF) based method \cite{mex2}. Both results are provided by \cite{shou2016action}. Comparison results are shown in Table \ref{result_mex}, our SSAD network achieve significant performance gain in all action categories of MEXaction2 and the mAP is increased from $7.4\%$ to $11.0\%$ with overlap threshold 0.5. Figure \ref{fig_vs} shows the visualization of prediction results for two action categories respectively.

\subsection{Model Analysis}

We evaluate SSAD network with different variants in THUMOS'14 to study their effects, including action classifiers, architectures of SSAD network and post-processing strategy.

\begin{table}[!tbp]
\centering
\caption{Comparisons between different action classifiers used in SSAD on THUMOS'14, where two-stream network includes both spatial and temporal networks.}
\begin{tabular}{p{5.7cm}<{\centering}p{2.1cm}<{\centering}}
\toprule
Action Classifier used for SAS Feature & mAP ($\theta =0.5$) \\
\midrule C3D Network  & 20.9 \\
	    Two-Stream Network & 21.9 \\
	   Two-Stream Network+C3D Network & \bf{24.6} \\
\bottomrule
\end{tabular}
\label{table_action}
\end{table}

\begin{table}[!tbp]
\centering
\caption{Comparisons among multiple base layers configurations on THUMOS'14. A, B, C, D, E are base layers configurations which presented in Figure \ref{fig_bn}.}
\begin{tabular}{p{3.8cm}<{\centering}p{0.5cm}<{\centering}p{0.5cm}<{\centering} p{0.5cm}<{\centering} p{0.5cm}<{\centering} p{0.5cm}<{\centering}}
\toprule
Network Configuration & A & B & C & D & E \\
\midrule mAP($\theta =0.5$)   & 23.7 & {\bf 24.6} & 24.1 & 23.9 & 23.4\\
\bottomrule
\end{tabular}
\label{table_base}
\end{table}

{\bf Action classifiers.}
Action classifiers are used to extract SAS feature. To study the contribution of different action classifiers, we evaluate them individually and coherently with  IoU threshold 0.5. As shown in Table \ref{table_action}, two-stream networks show better performance than C3D network and the combination of two-stream and C3D network lead to the best performance. In action recognition task such as UCF101, two-stream network \cite{wang2015towards} achieve $91.4\%$, which is better than $85.2\%$ of C3D \cite{tran2015learning} network (without combining with other method such as iDT \cite{wang2013action}). So two-stream network can predict action categories more precisely than C3D in snippet-level, which leads to a better performance of the SSAD network. Furthermore, the SAS feature extracted by two-stream network and C3D network are complementary and can achieve better result if used together.


{\bf Architectures of SSAD network.}
In section 3.3, we discuss several architectures used for base network of SSAD. These architectures have same input and output size. So we can evaluate them fairly without other changes of SSAD. The comparison results are shown in Table \ref{table_base}. Architecture $B$ achieves best performance among these configurations and is adopted for SSAD network. We can draw two conclusions from these results: (1) it is better to use max pooling layer instead of temporal convolutional layer to shorten the length of feature map; (2)  convolutional layers with kernel size 9 have better performance than other sizes.

\begin{table}[!tbp]
\centering
\caption{Evaluation on different post-processing strategy on THUMOS'14.}
\begin{tabular}{p{2cm}<{\centering} p{0.65cm}<{\centering} p{0.65cm}<{\centering} p{0.65cm}<{\centering} p{0.65cm}<{\centering} p{0.65cm}<{\centering} p{0.65cm}<{\centering}}
\toprule
$\bm{p_{class}}$ & \Checkmark & & \Checkmark & & \Checkmark & \Checkmark \\
$\bm{p_{sas}}$ &  & \Checkmark & \Checkmark &\Checkmark &  & \Checkmark \\
$p_{over}$ &  & &  &\Checkmark & \Checkmark & \Checkmark \\
\midrule
mAP ($\theta =0.5$) & 22.8 & 13.4 & 24.3 & 19.8 & 23.3 & {\bf 24.6} \\
\bottomrule
\end{tabular}
\label{table_post}

\end{table}

{\bf Post-processing strategy.} 
We evaluate multiple post-processing strategies. These strategies differ in the way of late fusion to generate $\bm{p_{final}}$ and are shown in Table \ref{table_post}.
For example, $\bm{p_{class}}$  is used for generate $\bm{p_{final}}$ if it is ticked in table.
Evaluation results are shown in Table \ref{table_post}. For the categories score, we can find that $\bm{p_{class}}$ has  better performance than $\bm{\bar{p}}_{sas}$. And using the multiplication factor $p_{over}$ can further improve the performance. SSAD network achieves the best performance with the complete post-processing strategy.

\section{Conclusion}
In this paper, we propose the Single Shot Action Detector (SSAD) network for temporal action detection task. Our SSAD network drops the proposal generation step and can directly predict action instances in untrimmed video. Also, we have explored many configurations of SSAD network to make SSAD network work better for temporal action detection. When setting Intersection-over-Union threshold to 0.5 during evaluation, SSAD significantly outperforms other state-of-the-art systems by increasing mAP from $19.0\%$ to $24.6\%$  on THUMOS'14 and from $7.4\%$ to $11.0\%$ on MEXaction2. In our approach, we conduct feature extraction and action detection separately, which makes SSAD network can handle concise high-level features and be easily trained. A promising future direction is to combine feature extraction procedure and SSAD network together to form an end-to-end framework, so that the whole framework can be trained from raw video directly.

\newpage

\bibliographystyle{ACM-Reference-Format}
\bibliography{sigproc} 


\begin{thebibliography}{00}


\ifx \showCODEN    \undefined \def \showCODEN     #1{\unskip}     \fi
\ifx \showDOI      \undefined \def \showDOI       #1{{\tt DOI:}\penalty0{#1}\ }
  \fi
\ifx \showISBNx    \undefined \def \showISBNx     #1{\unskip}     \fi
\ifx \showISBNxiii \undefined \def \showISBNxiii  #1{\unskip}     \fi
\ifx \showISSN     \undefined \def \showISSN      #1{\unskip}     \fi
\ifx \showLCCN     \undefined \def \showLCCN      #1{\unskip}     \fi
\ifx \shownote     \undefined \def \shownote      #1{#1}          \fi
\ifx \showarticletitle \undefined \def \showarticletitle #1{#1}   \fi
\ifx \showURL      \undefined \def \showURL       #1{#1}          \fi
\providecommand\bibfield[2]{#2}
\providecommand\bibinfo[2]{#2}
\providecommand\natexlab[1]{#1}
\providecommand\showeprint[2][]{arXiv:#2}

\bibitem[\protect\citeauthoryear{??}{mex}{2015}]%
        {mex2}
 \bibinfo{year}{2015}\natexlab{}.
\newblock \bibinfo{title}{MEXaction2}.
\newblock
  \bibinfo{howpublished}{\url{http://mexculture.cnam.fr/xwiki/bin/view/Datasets/Mex+action+dataset}}.
    (\bibinfo{year}{2015}).
\newblock


\bibitem[\protect\citeauthoryear{Abadi, Agarwal, Barham, et~al\mbox{.}}{Abadi
  et~al\mbox{.}}{2016}]%
        {abadi2016tensorflow}
\bibfield{author}{\bibinfo{person}{M. Abadi}, \bibinfo{person}{A. Agarwal},
  \bibinfo{person}{P. Barham}, {and} \bibinfo{person}{others}.}
  \bibinfo{year}{2016}\natexlab{}.
\newblock \showarticletitle{Tensorflow: Large-scale machine learning on
  heterogeneous distributed systems}.
\newblock \bibinfo{journal}{{\em arXiv preprint arXiv:1603.04467\/}}
  (\bibinfo{year}{2016}).
\newblock


\bibitem[\protect\citeauthoryear{Caba~Heilbron, Carlos~Niebles, and
  Ghanem}{Caba~Heilbron et~al\mbox{.}}{2016}]%
        {fast_temporal_activity_cvpr16}
\bibfield{author}{\bibinfo{person}{F. Caba~Heilbron}, \bibinfo{person}{J.
  Carlos~Niebles}, {and} \bibinfo{person}{B. Ghanem}.}
  \bibinfo{year}{2016}\natexlab{}.
\newblock \showarticletitle{Fast temporal activity proposals for efficient
  detection of human actions in untrimmed videos}. In \bibinfo{booktitle}{{\em
  Proceedings of the IEEE Conference on Computer Vision and Pattern
  Recognition}}. \bibinfo{pages}{1914--1923}.
\newblock


\bibitem[\protect\citeauthoryear{Deng, Dong, Socher, Li, Li, and Feifei}{Deng
  et~al\mbox{.}}{2009}]%
        {deng2009imagenet}
\bibfield{author}{\bibinfo{person}{J. Deng}, \bibinfo{person}{W. Dong},
  \bibinfo{person}{R. Socher}, \bibinfo{person}{L. Li}, \bibinfo{person}{K.
  Li}, {and} \bibinfo{person}{L. Feifei}.} \bibinfo{year}{2009}\natexlab{}.
\newblock \showarticletitle{ImageNet: A large-scale hierarchical image
  database}.
\newblock  (\bibinfo{year}{2009}), \bibinfo{pages}{248--255}.
\newblock


\bibitem[\protect\citeauthoryear{Escorcia, Heilbron, Niebles, and
  Ghanem}{Escorcia et~al\mbox{.}}{2016}]%
        {escorcia2016daps}
\bibfield{author}{\bibinfo{person}{V. Escorcia}, \bibinfo{person}{F.~C.
  Heilbron}, \bibinfo{person}{J.~C. Niebles}, {and} \bibinfo{person}{B.
  Ghanem}.} \bibinfo{year}{2016}\natexlab{}.
\newblock \showarticletitle{Daps: Deep action proposals for action
  understanding}. In \bibinfo{booktitle}{{\em European Conference on Computer
  Vision}}. Springer, \bibinfo{pages}{768--784}.
\newblock


\bibitem[\protect\citeauthoryear{Feichtenhofer, Pinz, and
  Zisserman}{Feichtenhofer et~al\mbox{.}}{2016}]%
        {feichtenhofer2016convolutional}
\bibfield{author}{\bibinfo{person}{C. Feichtenhofer}, \bibinfo{person}{A.
  Pinz}, {and} \bibinfo{person}{A. Zisserman}.}
  \bibinfo{year}{2016}\natexlab{}.
\newblock \showarticletitle{Convolutional two-stream network fusion for video
  action recognition}. In \bibinfo{booktitle}{{\em Proceedings of the IEEE
  Conference on Computer Vision and Pattern Recognition}}.
  \bibinfo{pages}{1933--1941}.
\newblock


\bibitem[\protect\citeauthoryear{Gemert, Jain, Gati, Snoek,
  et~al\mbox{.}}{Gemert et~al\mbox{.}}{2015}]%
        {apt}
\bibfield{author}{\bibinfo{person}{J. Gemert}, \bibinfo{person}{M. Jain},
  \bibinfo{person}{E. Gati}, \bibinfo{person}{C.~G. Snoek}, {and}
  \bibinfo{person}{others}.} \bibinfo{year}{2015}\natexlab{}.
\newblock \bibinfo{booktitle}{{\em Apt: Action localization proposals from
  dense trajectories}}.
\newblock \bibinfo{publisher}{BMVA Press}.
\newblock


\bibitem[\protect\citeauthoryear{Girshick}{Girshick}{2015}]%
        {girshick2015fastt}
\bibfield{author}{\bibinfo{person}{R. Girshick}.}
  \bibinfo{year}{2015}\natexlab{}.
\newblock \showarticletitle{Fast r-cnn}. In \bibinfo{booktitle}{{\em
  Proceedings of the IEEE International Conference on Computer Vision}}.
  \bibinfo{pages}{1440--1448}.
\newblock


\bibitem[\protect\citeauthoryear{Girshick, Donahue, Darrell, and
  Malik}{Girshick et~al\mbox{.}}{2014}]%
        {girshick2014rich}
\bibfield{author}{\bibinfo{person}{R. Girshick}, \bibinfo{person}{J. Donahue},
  \bibinfo{person}{T. Darrell}, {and} \bibinfo{person}{J. Malik}.}
  \bibinfo{year}{2014}\natexlab{}.
\newblock \showarticletitle{Rich feature hierarchies for accurate object
  detection and semantic segmentation}. In \bibinfo{booktitle}{{\em Proceedings
  of the IEEE conference on computer vision and pattern recognition}}.
  \bibinfo{pages}{580--587}.
\newblock


\bibitem[\protect\citeauthoryear{Glorot and Bengio}{Glorot and Bengio}{2010}]%
        {glorot2010understanding}
\bibfield{author}{\bibinfo{person}{X. Glorot} {and} \bibinfo{person}{Y.
  Bengio}.} \bibinfo{year}{2010}\natexlab{}.
\newblock \showarticletitle{Understanding the difficulty of training deep
  feedforward neural networks.}. In \bibinfo{booktitle}{{\em Aistats}},
  Vol.~\bibinfo{volume}{9}. \bibinfo{pages}{249--256}.
\newblock


\bibitem[\protect\citeauthoryear{Glorot, Bordes, and Bengio}{Glorot
  et~al\mbox{.}}{2011}]%
        {glorot2011deep}
\bibfield{author}{\bibinfo{person}{X. Glorot}, \bibinfo{person}{A. Bordes},
  {and} \bibinfo{person}{Y. Bengio}.} \bibinfo{year}{2011}\natexlab{}.
\newblock \showarticletitle{Deep Sparse Rectifier Neural Networks.}. In
  \bibinfo{booktitle}{{\em Aistats}}, Vol.~\bibinfo{volume}{15}.
  \bibinfo{pages}{275}.
\newblock


\bibitem[\protect\citeauthoryear{He, Zhang, Ren, and Sun}{He
  et~al\mbox{.}}{2016}]%
        {He_2016_CVPR}
\bibfield{author}{\bibinfo{person}{K. He}, \bibinfo{person}{X. Zhang},
  \bibinfo{person}{S. Ren}, {and} \bibinfo{person}{J. Sun}.}
  \bibinfo{year}{2016}\natexlab{}.
\newblock \showarticletitle{Deep residual learning for image recognition}. In
  \bibinfo{booktitle}{{\em Proceedings of the IEEE Conference on Computer
  Vision and Pattern Recognition}}. \bibinfo{pages}{770--778}.
\newblock


\bibitem[\protect\citeauthoryear{Jia, Shelhamer, Donahue, Karayev, Long,
  Girshick, Guadarrama, and Darrell}{Jia et~al\mbox{.}}{2014}]%
        {jia2014caffe}
\bibfield{author}{\bibinfo{person}{Y. Jia}, \bibinfo{person}{E. Shelhamer},
  \bibinfo{person}{J. Donahue}, \bibinfo{person}{S. Karayev},
  \bibinfo{person}{J. Long}, \bibinfo{person}{R. Girshick}, \bibinfo{person}{S.
  Guadarrama}, {and} \bibinfo{person}{T. Darrell}.}
  \bibinfo{year}{2014}\natexlab{}.
\newblock \showarticletitle{Caffe: Convolutional architecture for fast feature
  embedding}. In \bibinfo{booktitle}{{\em Proceedings of the 22nd ACM
  international conference on Multimedia}}. ACM, \bibinfo{pages}{675--678}.
\newblock


\bibitem[\protect\citeauthoryear{Jiang, Liu, Zamir, Toderici, Laptev, Shah, and
  Sukthankar}{Jiang et~al\mbox{.}}{2014}]%
        {jiang2014thumos}
\bibfield{author}{\bibinfo{person}{Y.~G. Jiang}, \bibinfo{person}{J. Liu},
  \bibinfo{person}{A.~R. Zamir}, \bibinfo{person}{G. Toderici},
  \bibinfo{person}{I. Laptev}, \bibinfo{person}{M. Shah}, {and}
  \bibinfo{person}{R. Sukthankar}.} \bibinfo{year}{2014}\natexlab{}.
\newblock \showarticletitle{THUMOS challenge: Action recognition with a large
  number of classes}. In \bibinfo{booktitle}{{\em ECCV Workshop}}.
\newblock


\bibitem[\protect\citeauthoryear{Karaman, Seidenari, and Del~Bimbo}{Karaman
  et~al\mbox{.}}{2014}]%
        {karaman2014fast}
\bibfield{author}{\bibinfo{person}{S. Karaman}, \bibinfo{person}{L. Seidenari},
  {and} \bibinfo{person}{A. Del~Bimbo}.} \bibinfo{year}{2014}\natexlab{}.
\newblock \showarticletitle{Fast saliency based pooling of fisher encoded dense
  trajectories}. In \bibinfo{booktitle}{{\em ECCV THUMOS Workshop}},
  Vol.~\bibinfo{volume}{1}.
\newblock


\bibitem[\protect\citeauthoryear{Karpathy, Toderici, Shetty, Leung, Sukthankar,
  and Fei-Fei}{Karpathy et~al\mbox{.}}{2014}]%
        {sports1m}
\bibfield{author}{\bibinfo{person}{A. Karpathy}, \bibinfo{person}{G. Toderici},
  \bibinfo{person}{S. Shetty}, \bibinfo{person}{T. Leung}, \bibinfo{person}{R.
  Sukthankar}, {and} \bibinfo{person}{L. Fei-Fei}.}
  \bibinfo{year}{2014}\natexlab{}.
\newblock \showarticletitle{Large-scale video classification with convolutional
  neural networks}. In \bibinfo{booktitle}{{\em Proceedings of the IEEE
  conference on Computer Vision and Pattern Recognition}}.
  \bibinfo{pages}{1725--1732}.
\newblock


\bibitem[\protect\citeauthoryear{Kingma and Ba}{Kingma and Ba}{2014}]%
        {kingma2014adam}
\bibfield{author}{\bibinfo{person}{D. Kingma} {and} \bibinfo{person}{J. Ba}.}
  \bibinfo{year}{2014}\natexlab{}.
\newblock \showarticletitle{Adam: A method for stochastic optimization}.
\newblock \bibinfo{journal}{{\em arXiv preprint arXiv:1412.6980\/}}
  (\bibinfo{year}{2014}).
\newblock


\bibitem[\protect\citeauthoryear{Kuehne, Jhuang, Stiefelhagen, and
  Serre}{Kuehne et~al\mbox{.}}{2013}]%
        {kuehne2013hmdb51}
\bibfield{author}{\bibinfo{person}{H. Kuehne}, \bibinfo{person}{H. Jhuang},
  \bibinfo{person}{R. Stiefelhagen}, {and} \bibinfo{person}{T. Serre}.}
  \bibinfo{year}{2013}\natexlab{}.
\newblock \showarticletitle{HMDB51: A large video database for human motion
  recognition}.
\newblock In \bibinfo{booktitle}{{\em High Performance Computing in Science and
  Engineering '12}}. \bibinfo{publisher}{Springer}, \bibinfo{pages}{571--582}.
\newblock


\bibitem[\protect\citeauthoryear{Lea, Vidal, Reiter, and Hager}{Lea
  et~al\mbox{.}}{2016}]%
        {lea2016temporal}
\bibfield{author}{\bibinfo{person}{C. Lea}, \bibinfo{person}{R. Vidal},
  \bibinfo{person}{A. Reiter}, {and} \bibinfo{person}{G.~D. Hager}.}
  \bibinfo{year}{2016}\natexlab{}.
\newblock \showarticletitle{Temporal Convolutional Networks: A Unified Approach
  to Action Segmentation}. In \bibinfo{booktitle}{{\em Computer Vision--ECCV
  2016 Workshops}}. Springer, \bibinfo{pages}{47--54}.
\newblock


\bibitem[\protect\citeauthoryear{Liu, Anguelov, Erhan, Szegedy, Reed, Fu, and
  Berg}{Liu et~al\mbox{.}}{2016}]%
        {liu2016ssd}
\bibfield{author}{\bibinfo{person}{W. Liu}, \bibinfo{person}{D. Anguelov},
  \bibinfo{person}{D. Erhan}, \bibinfo{person}{C. Szegedy}, \bibinfo{person}{S.
  Reed}, \bibinfo{person}{C. Fu}, {and} \bibinfo{person}{A.~C. Berg}.}
  \bibinfo{year}{2016}\natexlab{}.
\newblock \showarticletitle{SSD: Single shot multibox detector}. In
  \bibinfo{booktitle}{{\em European Conference on Computer Vision}}. Springer,
  \bibinfo{pages}{21--37}.
\newblock


\bibitem[\protect\citeauthoryear{Ma, Sigal, and Sclaroff}{Ma
  et~al\mbox{.}}{2016}]%
        {ma2016learning}
\bibfield{author}{\bibinfo{person}{S. Ma}, \bibinfo{person}{L. Sigal}, {and}
  \bibinfo{person}{S. Sclaroff}.} \bibinfo{year}{2016}\natexlab{}.
\newblock \showarticletitle{Learning activity progression in LSTMs for activity
  detection and early detection}. In \bibinfo{booktitle}{{\em Proceedings of
  the IEEE Conference on Computer Vision and Pattern Recognition}}.
  \bibinfo{pages}{1942--1950}.
\newblock


\bibitem[\protect\citeauthoryear{Mettes, van Gemert, and Snoek}{Mettes
  et~al\mbox{.}}{2016}]%
        {spoton_eccv16}
\bibfield{author}{\bibinfo{person}{P. Mettes}, \bibinfo{person}{J.~C. van
  Gemert}, {and} \bibinfo{person}{C.~G. Snoek}.}
  \bibinfo{year}{2016}\natexlab{}.
\newblock \showarticletitle{Spot on: Action localization from
  pointly-supervised proposals}. In \bibinfo{booktitle}{{\em European
  Conference on Computer Vision}}. Springer, \bibinfo{pages}{437--453}.
\newblock


\bibitem[\protect\citeauthoryear{Oneata, Verbeek, and Schmid}{Oneata
  et~al\mbox{.}}{2014}]%
        {oneata2014lear}
\bibfield{author}{\bibinfo{person}{D. Oneata}, \bibinfo{person}{J. Verbeek},
  {and} \bibinfo{person}{C. Schmid}.} \bibinfo{year}{2014}\natexlab{}.
\newblock \showarticletitle{The LEAR submission at Thumos 2014}.
\newblock \bibinfo{journal}{{\em ECCV THUMOS Workshop\/}}
  (\bibinfo{year}{2014}).
\newblock


\bibitem[\protect\citeauthoryear{Qiu, Yao, and Mei}{Qiu et~al\mbox{.}}{2016}]%
        {qiu2016deep}
\bibfield{author}{\bibinfo{person}{Z. Qiu}, \bibinfo{person}{T. Yao}, {and}
  \bibinfo{person}{T. Mei}.} \bibinfo{year}{2016}\natexlab{}.
\newblock \showarticletitle{Deep Quantization: Encoding Convolutional
  Activations with Deep Generative Model}.
\newblock \bibinfo{journal}{{\em arXiv preprint arXiv:1611.09502\/}}
  (\bibinfo{year}{2016}).
\newblock


\bibitem[\protect\citeauthoryear{Redmon, Divvala, Girshick, and Farhadi}{Redmon
  et~al\mbox{.}}{2016}]%
        {redmon2016you}
\bibfield{author}{\bibinfo{person}{J. Redmon}, \bibinfo{person}{S. Divvala},
  \bibinfo{person}{R. Girshick}, {and} \bibinfo{person}{A. Farhadi}.}
  \bibinfo{year}{2016}\natexlab{}.
\newblock \showarticletitle{You only look once: Unified, real-time object
  detection}. In \bibinfo{booktitle}{{\em Proceedings of the IEEE Conference on
  Computer Vision and Pattern Recognition}}. \bibinfo{pages}{779--788}.
\newblock


\bibitem[\protect\citeauthoryear{Redmon and Farhadi}{Redmon and
  Farhadi}{2016}]%
        {redmon2016yolo9000}
\bibfield{author}{\bibinfo{person}{J. Redmon} {and} \bibinfo{person}{A.
  Farhadi}.} \bibinfo{year}{2016}\natexlab{}.
\newblock \showarticletitle{YOLO9000: Better, Faster, Stronger}.
\newblock \bibinfo{journal}{{\em arXiv preprint arXiv:1612.08242\/}}
  (\bibinfo{year}{2016}).
\newblock


\bibitem[\protect\citeauthoryear{Ren, He, Girshick, and Sun}{Ren
  et~al\mbox{.}}{2015}]%
        {ren2015faster}
\bibfield{author}{\bibinfo{person}{S. Ren}, \bibinfo{person}{K. He},
  \bibinfo{person}{R. Girshick}, {and} \bibinfo{person}{J. Sun}.}
  \bibinfo{year}{2015}\natexlab{}.
\newblock \showarticletitle{Faster r-cnn: Towards real-time object detection
  with region proposal networks}. In \bibinfo{booktitle}{{\em Advances in
  neural information processing systems}}. \bibinfo{pages}{91--99}.
\newblock


\bibitem[\protect\citeauthoryear{Richard and Gall}{Richard and Gall}{2016}]%
        {Richard_2016_CVPR}
\bibfield{author}{\bibinfo{person}{A. Richard} {and} \bibinfo{person}{J.
  Gall}.} \bibinfo{year}{2016}\natexlab{}.
\newblock \showarticletitle{Temporal action detection using a statistical
  language model}. In \bibinfo{booktitle}{{\em Proceedings of the IEEE
  Conference on Computer Vision and Pattern Recognition}}.
  \bibinfo{pages}{3131--3140}.
\newblock


\bibitem[\protect\citeauthoryear{Shou, Wang, and Chang}{Shou
  et~al\mbox{.}}{2016}]%
        {shou2016action}
\bibfield{author}{\bibinfo{person}{Z. Shou}, \bibinfo{person}{D. Wang}, {and}
  \bibinfo{person}{S.-F. Chang}.} \bibinfo{year}{2016}\natexlab{}.
\newblock \showarticletitle{Temporal action localization in untrimmed videos
  via multi-stage cnns}. In \bibinfo{booktitle}{{\em Proceedings of the IEEE
  Conference on Computer Vision and Pattern Recognition}}.
  \bibinfo{pages}{1049--1058}.
\newblock


\bibitem[\protect\citeauthoryear{Simonyan and Zisserman}{Simonyan and
  Zisserman}{2014}]%
        {simonyan2014two}
\bibfield{author}{\bibinfo{person}{K. Simonyan} {and} \bibinfo{person}{A.
  Zisserman}.} \bibinfo{year}{2014}\natexlab{}.
\newblock \showarticletitle{Two-stream convolutional networks for action
  recognition in videos}. In \bibinfo{booktitle}{{\em Advances in Neural
  Information Processing Systems}}. \bibinfo{pages}{568--576}.
\newblock


\bibitem[\protect\citeauthoryear{Simonyan and Zisserman}{Simonyan and
  Zisserman}{2015}]%
        {Simonyan15}
\bibfield{author}{\bibinfo{person}{K. Simonyan} {and} \bibinfo{person}{A.
  Zisserman}.} \bibinfo{year}{2015}\natexlab{}.
\newblock \showarticletitle{Very Deep Convolutional Networks for Large-Scale
  Image Recognition}. In \bibinfo{booktitle}{{\em International Conference on
  Learning Representations}}.
\newblock


\bibitem[\protect\citeauthoryear{Singh, Marks, Jones, Tuzel, and Shao}{Singh
  et~al\mbox{.}}{2016}]%
        {singh2016multi}
\bibfield{author}{\bibinfo{person}{B. Singh}, \bibinfo{person}{T.~K. Marks},
  \bibinfo{person}{M. Jones}, \bibinfo{person}{O. Tuzel}, {and}
  \bibinfo{person}{M. Shao}.} \bibinfo{year}{2016}\natexlab{}.
\newblock \showarticletitle{A multi-stream bi-directional recurrent neural
  network for fine-grained action detection}. In \bibinfo{booktitle}{{\em
  Proceedings of the IEEE Conference on Computer Vision and Pattern
  Recognition}}. \bibinfo{pages}{1961--1970}.
\newblock


\bibitem[\protect\citeauthoryear{Singh and Cuzzolin}{Singh and
  Cuzzolin}{2016}]%
        {singh2016untrimmed}
\bibfield{author}{\bibinfo{person}{G. Singh} {and} \bibinfo{person}{F.
  Cuzzolin}.} \bibinfo{year}{2016}\natexlab{}.
\newblock \showarticletitle{Untrimmed Video Classification for Activity
  Detection: submission to ActivityNet Challenge}.
\newblock \bibinfo{journal}{{\em arXiv preprint arXiv:1607.01979\/}}
  (\bibinfo{year}{2016}).
\newblock


\bibitem[\protect\citeauthoryear{Soomro, Zamir, and Shah}{Soomro
  et~al\mbox{.}}{2012}]%
        {soomro2012ucf101}
\bibfield{author}{\bibinfo{person}{K. Soomro}, \bibinfo{person}{A.~R. Zamir},
  {and} \bibinfo{person}{M. Shah}.} \bibinfo{year}{2012}\natexlab{}.
\newblock \showarticletitle{UCF101: A dataset of 101 human actions classes from
  videos in the wild}.
\newblock \bibinfo{journal}{{\em arXiv preprint arXiv:1212.0402\/}}
  (\bibinfo{year}{2012}).
\newblock


\bibitem[\protect\citeauthoryear{Szegedy, Liu, Jia, Sermanet, Reed, Anguelov,
  Erhan, Vanhoucke, and Rabinovich}{Szegedy et~al\mbox{.}}{2015}]%
        {szegedy2015going}
\bibfield{author}{\bibinfo{person}{C. Szegedy}, \bibinfo{person}{W. Liu},
  \bibinfo{person}{Y. Jia}, \bibinfo{person}{P. Sermanet}, \bibinfo{person}{S.
  Reed}, \bibinfo{person}{D. Anguelov}, \bibinfo{person}{D. Erhan},
  \bibinfo{person}{V. Vanhoucke}, {and} \bibinfo{person}{A. Rabinovich}.}
  \bibinfo{year}{2015}\natexlab{}.
\newblock \showarticletitle{Going deeper with convolutions}. In
  \bibinfo{booktitle}{{\em Proceedings of the IEEE Conference on Computer
  Vision and Pattern Recognition}}. \bibinfo{pages}{1--9}.
\newblock


\bibitem[\protect\citeauthoryear{Tran, Bourdev, Fergus, Torresani, and
  Paluri}{Tran et~al\mbox{.}}{2015}]%
        {tran2015learning}
\bibfield{author}{\bibinfo{person}{D. Tran}, \bibinfo{person}{L. Bourdev},
  \bibinfo{person}{R. Fergus}, \bibinfo{person}{L. Torresani}, {and}
  \bibinfo{person}{M. Paluri}.} \bibinfo{year}{2015}\natexlab{}.
\newblock \showarticletitle{Learning spatiotemporal features with 3d
  convolutional networks}. In \bibinfo{booktitle}{{\em Proceedings of the IEEE
  International Conference on Computer Vision}}. \bibinfo{pages}{4489--4497}.
\newblock


\bibitem[\protect\citeauthoryear{Wang, Kl{\"a}ser, Schmid, and Liu}{Wang
  et~al\mbox{.}}{2011}]%
        {dtf}
\bibfield{author}{\bibinfo{person}{H. Wang}, \bibinfo{person}{A. Kl{\"a}ser},
  \bibinfo{person}{C. Schmid}, {and} \bibinfo{person}{C.-L. Liu}.}
  \bibinfo{year}{2011}\natexlab{}.
\newblock \showarticletitle{Action recognition by dense trajectories}. In
  \bibinfo{booktitle}{{\em Computer Vision and Pattern Recognition (CVPR), 2011
  IEEE Conference on}}. IEEE, \bibinfo{pages}{3169--3176}.
\newblock


\bibitem[\protect\citeauthoryear{Wang and Schmid}{Wang and Schmid}{2013}]%
        {wang2013action}
\bibfield{author}{\bibinfo{person}{H. Wang} {and} \bibinfo{person}{C. Schmid}.}
  \bibinfo{year}{2013}\natexlab{}.
\newblock \showarticletitle{Action recognition with improved trajectories}. In
  \bibinfo{booktitle}{{\em Proceedings of the IEEE International Conference on
  Computer Vision}}. \bibinfo{pages}{3551--3558}.
\newblock


\bibitem[\protect\citeauthoryear{Wang, Qiao, and Tang}{Wang
  et~al\mbox{.}}{2014}]%
        {wang2014action}
\bibfield{author}{\bibinfo{person}{L. Wang}, \bibinfo{person}{Y. Qiao}, {and}
  \bibinfo{person}{X. Tang}.} \bibinfo{year}{2014}\natexlab{}.
\newblock \showarticletitle{Action recognition and detection by combining
  motion and appearance features}.
\newblock \bibinfo{journal}{{\em THUMOS14 Action Recognition Challenge\/}}
  \bibinfo{volume}{1} (\bibinfo{year}{2014}), \bibinfo{pages}{2}.
\newblock


\bibitem[\protect\citeauthoryear{Wang, Xiong, Wang, and Qiao}{Wang
  et~al\mbox{.}}{2015}]%
        {wang2015towards}
\bibfield{author}{\bibinfo{person}{L. Wang}, \bibinfo{person}{Y. Xiong},
  \bibinfo{person}{Z. Wang}, {and} \bibinfo{person}{Y. Qiao}.}
  \bibinfo{year}{2015}\natexlab{}.
\newblock \showarticletitle{Towards good practices for very deep two-stream
  convnets}.
\newblock \bibinfo{journal}{{\em arXiv preprint arXiv:1507.02159\/}}
  (\bibinfo{year}{2015}).
\newblock


\bibitem[\protect\citeauthoryear{Wang and Tao}{Wang and Tao}{2016}]%
        {wang2016uts}
\bibfield{author}{\bibinfo{person}{R. Wang} {and} \bibinfo{person}{D. Tao}.}
  \bibinfo{year}{2016}\natexlab{}.
\newblock \showarticletitle{UTS at activitynet 2016}.
\newblock \bibinfo{journal}{{\em AcitivityNet Large Scale Activity Recognition
  Challenge\/}}  \bibinfo{volume}{2016} (\bibinfo{year}{2016}),
  \bibinfo{pages}{8}.
\newblock


\bibitem[\protect\citeauthoryear{Yeung, Russakovsky, Mori, and Fei-Fei}{Yeung
  et~al\mbox{.}}{2016}]%
        {yeung2015end}
\bibfield{author}{\bibinfo{person}{S. Yeung}, \bibinfo{person}{O. Russakovsky},
  \bibinfo{person}{G. Mori}, {and} \bibinfo{person}{L. Fei-Fei}.}
  \bibinfo{year}{2016}\natexlab{}.
\newblock \showarticletitle{End-to-end learning of action detection from frame
  glimpses in videos}. In \bibinfo{booktitle}{{\em Proceedings of the IEEE
  Conference on Computer Vision and Pattern Recognition}}.
  \bibinfo{pages}{2678--2687}.
\newblock


\bibitem[\protect\citeauthoryear{Yu and Yuan}{Yu and Yuan}{2015}]%
        {gangyu}
\bibfield{author}{\bibinfo{person}{G. Yu} {and} \bibinfo{person}{J. Yuan}.}
  \bibinfo{year}{2015}\natexlab{}.
\newblock \showarticletitle{Fast action proposals for human action detection
  and search}. In \bibinfo{booktitle}{{\em Proceedings of the IEEE Conference
  on Computer Vision and Pattern Recognition}}. \bibinfo{pages}{1302--1311}.
\newblock


\bibitem[\protect\citeauthoryear{Yuan, Ni, Yang, and Kassim}{Yuan
  et~al\mbox{.}}{2016}]%
        {Yuan2016Temporal}
\bibfield{author}{\bibinfo{person}{J. Yuan}, \bibinfo{person}{B. Ni},
  \bibinfo{person}{X. Yang}, {and} \bibinfo{person}{A.~A. Kassim}.}
  \bibinfo{year}{2016}\natexlab{}.
\newblock \showarticletitle{Temporal Action Localization with Pyramid of Score
  Distribution Features}. In \bibinfo{booktitle}{{\em IEEE Conference on
  Computer Vision and Pattern Recognition}}. \bibinfo{pages}{3093--3102}.
\newblock


\bibitem[\protect\citeauthoryear{Zhu and Newsam}{Zhu and Newsam}{2016}]%
        {zhu2016efficient}
\bibfield{author}{\bibinfo{person}{Y. Zhu} {and} \bibinfo{person}{S. Newsam}.}
  \bibinfo{year}{2016}\natexlab{}.
\newblock \showarticletitle{Efficient Action Detection in Untrimmed Videos via
  Multi-Task Learning}.
\newblock \bibinfo{journal}{{\em arXiv preprint arXiv:1612.07403\/}}
  (\bibinfo{year}{2016}).
\newblock


\end{thebibliography}

\end{document}